# SEAL: Semisupervised Adversarial Active Learning on Attributed Graphs

Yayong Li, Jie Yin, *Member, IEEE*, and Ling Chen, *Senior Member, IEEE*

***Abstract*—Active learning (AL) on attributed graphs has received increasing attention with the prevalence of graph-structured data. Although AL has been widely studied for alleviating label sparsity issues with the conventional nonrelational data, how to make it effective over attributed graphs remains an open research question. Existing AL algorithms on node classification attempt to reuse the classic AL query strategies designed for nonrelational data. However, they suffer from two major limitations. First, different AL query strategies calculated in distinct scoring spaces are often naively combined to determine which nodes to be labeled. Second, the AL query engine and the learning of the classifier are treated as two separating processes, resulting in unsatisfactory performance. In this article, we propose a SEmisupervised Adversarial active Learning (SEAL) framework on attributed graphs, which fully leverages the representation power of deep neural networks and devises a novel AL query strategy for node classification in an adversarial way. Our framework learns two adversarial components; a graph embedding network that encodes both the unlabeled and labeled nodes into a common latent space, expecting to trick the discriminator to regard all nodes as already labeled, and a semisupervised discriminator network that distinguishes the unlabeled from the existing labeled nodes. The divergence score, generated by the discriminator in a unified latent space, serves as the informativeness measure to actively select the most informative node to be labeled by an oracle. The two adversarial components form a closed loop to mutually and simultaneously reinforce each other toward enhancing the AL performance. Extensive experiments on real-world networks validate the effectiveness of the SEAL framework with superior performance improvements to state-of-the-art baselines on node classification tasks.**

## I. Introduction

RECENT years have witnessed an enormous growth of content-rich network data, in forms of social networks, citation networks, financial networks, and so on, generated from various domains. To enable effective knowledge discovery, graphs have become a powerful tool to represent network data, where nodes denote instances (e.g., users or documents) that are often characterized by rich content features, and edges denote relationships or interactions between nodes (e.g., friendship or citation relationships). Node classification is one of the most important tasks in analyzing such content-rich networks, which aims to predict the labels of unlabeled nodes given a partially labeled network. Classic methods such as the iterative classification algorithm (ICA) and recently proposed network embedding-based methods and graph neural networks [1] have been demonstrated to be effective in classifying many real-world networks. These algorithms rely on a sufficient number of labeled nodes provided to ensure desirable classification accuracy. Very often, however, acquiring a large quantity of node labels requires expert efforts and is very costly and time-consuming, which significantly limits the true success of these algorithms [2], [3].

In response, active learning (AL) has been proposed to alleviate the label sparsity issue in classifying sparsely labeled networks. It aims at maximizing the learning ability of classification models with the least labeling costs. An AL framework consists of two primary components: a query engine that selects an instance from the unlabeled pool to query its label and an oracle that provides a label to the queried instance. Different AL algorithms have been proposed in the last decade, which measure the informativeness of instances with certain criteria and selectively label instances that most potentially lead to performance improvements. For example, Lewis and Catlett [4] chosen to label the instances for which the classifier predicts with the least confidence, and Roy and McCallum [5] proposed to label the instances that are most likely to bring the model error reduction on the unlabeled pool. These active instance selection strategies allow constructing a more accurate model with fewer labels. However, when AL meets graph-structured data, these classic AL strategies that are effective for nonrelational data, such as query-by-committee [6], [7] or uncertainty sampling [4], [8], fail to achieve satisfactory results because of their inability to exploit the topological structure of graphs. Accordingly, a series of graph-aware AL strategies have been proposed [9]–[11]. These methods directly minimize the expected generalization error or variance of the classifiers and often incur high computational costs. They are designed based on graph structure only and thus lack the ability to exert the already labeled nodes and rich node features that indeed provide leverage to node classification. By contrast, ICA-based algorithms [12]–[14] leverage label dependence among neighboring nodes to choose the most informative nodes that can best improve the classifier built on the original node feature space and thus have limited classification performance.

Manuscript received July 22, 2019; revised March 16, 2020; accepted July 4, 2020. This work was supported in part by the USYD-Data61 Collaborative Research Project grant, in part by the Australian Research Council under Grant DP180100966, and in part by the China Scholarship Council under Grant 201806070131. *(Corresponding author: Jie Yin.)*

Yayong Li and Ling Chen are with the Faculty of Engineering and Information Technology, University of Technology Sydney, Sydney, NSW 2007, Australia (e-mail: yayong.li@student.uts.edu.au; ling.chen@uts.edu.au).

Jie Yin is with the Discipline of Business Analytics, The University of Sydney, Sydney, NSW 2006, Australia (e-mail: jie.yin@sydney.edu.au).

Color versions of one or more of the figures in this article are available online at http://ieeexplore.ieee.org.

Recently, graph neural networks (GNNs) [1], [15], [16] have achieved remarkable success by exploiting the power of deep learning in dealing with graph-structured data. Compared with traditional methods such as ICA [17] or Deep Walk-based embedding methods [18], GNN offers its notable advantages through representation learning in terms of capturing graph structure and aggregating neighboring information. Thus, GNN and its variants have achieved state-of-the-art results on both node and graph classification tasks. Yet, research on how to leverage GNN to empower AL on graphs is largely unexplored. AGE [19] and ANRMAB [20] are two AL algorithms that attempt to integrate graph convolutional networks (GCNs) with three AL query strategies, namely graph centrality, information density, and information entropy. While the former uses a linear combination of these strategies, the latter utilizes a multiarmed bandit (MAB) mechanism to dynamically adjust the weights on the respective strategies according to the MAB reward. However, the two algorithms share common weaknesses. First, they use a naive combination of three AL criteria as the informativeness measure. The AL criteria combined still operate separately on different scoring spaces, failing to capture interaction and interrelatedness between different factors. Second, they use the output of a GCN in a postprocessing way to determine the AL query strategy. This means that the AL query engine and the learning of graph embedding still work as two separating pieces; although the newly labeled node can improve the learning of GCN, the capacity of the query engine remains unchanged, leading to unsatisfactory performance improvements.

In this article, we propose a novel SEmisupervised Adversarial active Learning (SEAL) framework that seamlessly integrates AL with deep neural networks to select the most informative nodes to label on attributed graphs. The proposed framework explicitly asserts the usefulness of unlabeled nodes with regard to the existing labeled data. Inspired by adversarial learning, we define an informativeness measure based on the intuition that the unlabeled nodes differing the most with the labeled ones carry the most auxiliary information on what the classifier desires the most. Our SEAL framework comprises two adversarial components, a graph embedding network and a semisupervised discriminator network, which form a closed loop to actively collaborate with each other. The graph embedding network is trained to embed both the labeled and unlabeled nodes into the same latent space that encodes both graph structure and node features, expecting to fool the discriminator to regard all nodes as already labeled. The discriminator learns how to differentiate the unlabeled from the already labeled nodes. Instead of using a binary discriminator, we design a semisupervised discriminator with multiple outputs. The outputs, on the one hand receiving supervision from the existing labels, serve as class predictions, and on the other hand, produce a unified informativeness score in a common latent space. This score measures the divergence between the unlabeled and already labeled nodes. The unlabeled node with the highest score is selected to query its label. At the same time, the loss of the discriminator is backpropagated to the graph embedding network. The two adversarial components mutually reinforce each other in an iterative way to boost the AL performance.

The main contribution of this article is threefold.

1) We propose a novel adversarial AL framework that seamlessly incorporates AL into GNNs. Unlike previous methods that simply combine AL strategies residing at different scoring spaces, SEAL generates a unified informativeness score in a common latent space to enable instance selection, rendering the most desirable performance gains.
2) To the best of our knowledge, we are the first to propose an Semisupervised Adversarial Learning (SAL) structure with multiple outputs for AL on attributed graphs. This offers an advantage that the graph embedding network and the discriminator can collaborate with each other to mutually strengthen their performance.
3) We validate our SEAL framework through extensive experiments and ablation studies on four real-world networks, demonstrating its superior performance to state-of-the-art baselines on node classification tasks.

The rest of this article is organized as follows. Section II reviews the related literature. The problem statement and preliminaries are given in Section III. We present our proposed framework in Section IV and report experimental results in Section V. Finally, we conclude this article in Section VI.

## II. Related Work

In this section, we review related work from two main branches of research studies, namely classic AL strategies and AL on graphs.

### A. Classic AL Strategies

AL is a machine learning framework that aims to reduce the labeling cost when learning a prediction model by selecting the most informative instances to label. In the past decade, a variety of AL algorithms have been proposed to optimize the training performance given a fixed labeling budget. These algorithms differ mostly in the query strategy that they use to specify the informativeness criterion when selecting the best instances to label. Depending on what query strategies are used, classic AL strategies can be grouped into six categories [27], [28]: uncertainty sampling [4], [8], query-by-committee (QBC) [6], [7], expected model change (EMC) [21], expected error reduction (EER) [5], [22], expected variance reduction (EVR) [23], [24], and density-weighted methods (DWMs) [25], [26]. The core ideas and detailed comparisons of these algorithms are summarized in Table I. In general, these classic AL strategies can be instrumented with different classification algorithms. Among others, expected error/variance reduction strategies tend to render better empirical results because they iterate over the entire unlabeled pool to directly optimize model performance. However, they suffer from high computational overhead.

These methods have been shown to achieve good performance on nonrelational data. However, they are not sufficiently effective for graph-structured data, where data dependence needs to be incorporated into the AL process.

TABLE I
COMPARISON OF DIFFERENT CLASSIC AL STRATEGIES

| Categories | Core idea | Advantages | Disadvantages |
| --- | --- | --- | --- |
| Uncertainty sampling [4], [8] | Query the instances whose labels are predicted with the least confidence based on the current labeled set | Simple and fast | Prone to selecting noisy or unrepresentative instances |
| Query-by-Committee (QBC) [6], [7] | Query the instances with which multiple classifiers most disagree | | |
| Expected Model Change (EMC) [21] | Query the instances which would result in the most change to the current model parameters in the gradient of objective function | Directly optimize the model performance | Applicable only to gradient-based training methods, and incur huge computational cost when the feature or label space are large |
| Expected Error Reduction (EER) [5], [22] | Query the instances which most likely reduce the largest generalization error on the unlabeled pool | Minimize generalization errors by considering the entire input space, eliminating the disturbance of outliers | High computational complexity |
| Expected Variance Reduction (EVR) [23], [24] | Query the instances which minimize model variance | | |
| Density-Weighted Methods (DWM) [25], [26] | Query the instances that are representative of the underlying distribution of training data | Able to avoid selecting noisy instances | Not informative enough, and often combined with other strategies |

### *B. Active Learning on Graphs*

In recent years, AL on graphs has attracted significant attention to alleviate the label sparsity issues on graphs. Early research has focused on using graph-based metrics (e.g., centrality and impact) to calculate the AL query scores when selecting the nodes to label [29]. Other attempts have been made to directly optimize an objective function over graphs, where graph structure is utilized to train a classifier, such as graph cut-based method [30], [31], Gaussian field and harmonic function (GFHF) [9], learning with local and global consistency (LLGC) [10], and label propagation (LP) [32]. These methods aim to minimize the expected generalization error or variance of the classifier built using a graph structure. However, they often suffer from high computational complexity and are difficult to scale up. To improve efficiency, Zhao *et al.* [33] proposed to narrow the search space by sampling structurally important nodes in advance, and Zhu *et al.* [34] used uncertainty and graph centrality to prune the candidate set. However, these methods have assessed the informativeness of unlabeled nodes using only graph structure, while rich node features have not been fully explored to best inform the design of AL query strategies.

Another line of research formulates AL query strategies by integrating node features with graph topological structure [12]–[14], [35], [36]. Most methods design an ICA classifier by combining graph structure with node-specific features and then use various AL query strategies for instance selection. For example, ALFNET [12] adopts clustering techniques to form an initial labeled set. At each iteration, ALFNET aggregates neighboring labels with original node features to train three classifiers and computes a local disagreement score for each node. The scores are then aggregated for each cluster and the clusters with the highest scores are chosen, from which a set of nodes are selected to label. These methods, however, have focused on improving the classifiers built in the original feature space, thus leading to suboptimal prediction accuracy compared with counterparts built in latent feature spaces by deep learning models.

Only recently, researchers have proposed to exploit deep learning to empower AL on graphs. In virtue of great representation power of GNN, AGE [19] and ANRMAB [20] propose to incorporate a GCN into traditional AL strategies, which achieves a significant improvement compared with the previous methods. Both methods combine three traditional query strategies: graph centrality, information density, and uncertainty sampling. AGE uses a naive linear combination of the three strategies, whereas ANRMAB further adopts an MAB mechanism to adjust the weights of different strategies. However, the combined AL query engine and the learning of graph embedding still work as two separate and independent processes, resulting in limited AL performance gains. To fill the gap, our work is proposed to fully integrate the learning of graph embedding with a novel AL query strategy via a semisupervised discriminator. The learning of graph embedding and discriminator function is two adversarial components, which collaborate with each other to mutually strengthen their performance toward better AL performance.

## III. PROBLEM STATEMENT AND PRELIMINARIES

This section gives a formal problem definition and reviews the preliminaries of GCN and adversarial learning.

### *A. Problem Statement*

Given an undirected attributed graph $\mathcal{G} = \{\mathcal{V}, \mathcal{E}, \mathcal{X}\}$ with its adjacent matrix $A \in \mathbb{R}^{N \times N}$, where $\mathcal{V}$ denotes the node set, $\mathcal{E} \subseteq \mathcal{V} \times \mathcal{V}$ denotes the edge set, $\mathcal{X} = [\mathbf{x}_1, \mathbf{x}_2, \ldots, \mathbf{x}_N]^T \in \mathcal{R}^{N \times M}$ denotes the node feature matrix, and $\mathbf{x}_i \in \mathcal{R}^M$ is the $M$-dimensional feature vector of node $v_i \in \mathcal{V}$. Let $L = \{(\mathbf{x}_i, \mathbf{y}_i)\}_{i=1}^{|L|}$ denote a set of labeled nodes, where $\mathbf{y}_i = \{y_{i1}, \ldots, y_{ij}, \ldots, y_{iK}\}$ is the one-hot encoding of node $v_i$'s class label, with $y_{ij} \in \{0, 1\}$ and $K$ being the number of classes. The rest of the nodes belong to the unlabeled set $U$.

This work focuses on pool-based AL for node classification. Given a pool of unlabeled nodes, it works sequentially to select one unlabeled node to label at a time while retraining the classifier at each iteration.

Given a fixed labeling budget $B$ and an initial labeled set $L$, the AL problem aims to design a good query strategy $Q(\mathbf{x}_i; \Theta)$ parameterized with $\Theta$ that specifies which unlabeled node should be selected to label at each iteration. $Q(\mathbf{x}_i; \Theta)$ can also be considered as a utility function that assigns a score to each unlabeled node indicating its informativeness to the current classifier. The unlabeled node that maximizes $Q(\mathbf{x}_i; \Theta)$ is selected to label so that the retrained classifier can achieve the maximum accuracy [37]. Formally, this can be expressed as follows:

$$\mathbf{x}^* = \underset{\mathbf{x}_i \in U}{\operatorname{argmax}}\, Q(\mathbf{x}_i; \Theta) \tag{1}$$

where $\mathbf{x}^*$ represents the most informative node that is selected to query its label at each iteration and $\Theta$ represents the parameters of the learned query strategy.

### B. Preliminaries on GCN

The SEAL framework leverages GNN's representation power to empower AL on graphs. It is generic in nature and can be directly applied to any other graph embedding algorithms. In this work, we adopt GCN [1], as an example graph embedding algorithm, for semisupervised node classification in our AL framework.

GCN directly encodes the graph structure using a multilayer neural network model. A two-layer GCN propagation rule of GCN is defined as

$$Z = f(A, \mathcal{X}) = \operatorname{softmax}(\hat{A}\sigma(\hat{A}\mathcal{X}W^{(0)})W^{(1)}) \quad \text{with } \hat{A} = \widetilde{D}^{-\frac{1}{2}}\widetilde{A}\widetilde{D}^{-\frac{1}{2}} \tag{2}$$

where $\widetilde{A} = A + I_N$ is the adjacent matrix with added self-connections. $I_N$ is the identity matrix and $\widetilde{D_{ii}} = \sum_j \widetilde{A}_{ij}$. $W^{(l)}$ is the layer-specific trainable weight matrix in the $l$th layer. $\sigma(\cdot)$ is the activation function. Finally, the supervised loss function is defined as the cross-entropy error over all labeled nodes as

$$J_{\text{GCN}} = -\sum_{l=1}^{|L|}\sum_{k=1}^{K} y_{lk} \log z_{lk} \tag{3}$$

where $\mathbf{z}_l = \{z_{l1}, z_{l2}, \ldots, z_{lK}\} \in Z$ is the class predictions of node $v_l$.

### C. Adversarial Learning

Generative adversarial networks (GANs) [38] have emerged as a powerful framework for learning deep representations of arbitrarily complex data distributions via an adversarial process. A regular GAN sets up an adversarial platform for a generator $g_\theta(\cdot)$ and a discriminator $d_\phi(\cdot)$, where $g_\theta(\cdot)$ intends to produce samples as close to the real data as possible, whereas $d_\phi(\cdot)$ tries to tell apart samples either from the real data ($\mathbf{x} \sim \mathcal{P}_{\text{data}}(\mathbf{x})$) or from the generator ($\mathbf{s} \sim \mathcal{P}_g(\mathbf{s})$) as accurately as possible. This adversarial process is formulated as a min–max game with the following loss function:

$$\min_\theta \max_\phi E_{\mathbf{x}\sim\mathcal{P}_{\text{data}}(\mathbf{x})} \log d_\phi(\mathbf{x}) + E_{\mathbf{s}\sim\mathcal{P}_g(\mathbf{s})} \log(1 - d_\phi(g_\theta(\mathbf{s}))). \tag{4}$$

The conventional GAN framework finally converges at the state where $g_\theta(\cdot)$ recovers the training data perfectly and $d_\phi(\cdot)$ predicts 0.5 everywhere. Recently, this adversarial idea is adopted to solve AL problems on sequence and image classification. Deng *et al.* [39] designed a sequence-based AL algorithm that utilizes a binary adversarial network to shrink the search space of candidate samples, while an uncertainty strategy has to be applied to determine which sample should be chosen. Sinha *et al.* [40] used a similar adversarial active learning method to discriminate labeled and unlabeled images. However, both methods use a fully supervised binary discriminator and ignore the original class probability distribution. Our work was developed independently of these works, which yet focuses on graph-structured data where graph topological structure needs to be properly exploited. Our adversarial discriminator works in a semisupervised manner, which allows to distinguish labeled from unlabeled nodes and to model the class probability distribution.

## IV. SEAL Framework

This section presents the overview of the SEAL framework, followed by a detailed description of the main components.

### A. Framework Overview

Our AL objective is to select the most informative nodes to be labeled so as to robustly improve the node classification performance with the minimal labeling cost. To this end, we incorporate the adversarial learning method into AL through integrating one graph embedding network $G(\cdot)$ with a semisupervised discriminator network $D(\cdot)$. $G(\cdot)$ embeds both the labeled and unlabeled nodes into a common latent space with the uniform distribution to maximally confuse the discriminator, whereas $D(\cdot)$ intends to distinguish the unlabeled from labeled data as much as possible. The discriminator $D(\cdot)$ is iteratively reinforced through this competitive process, which provides a unified quantitative criterion to measure the divergence of the unlabeled nodes with respect to the existing labeled nodes. This criterion ideally enables the selection of the most informative unlabeled node to be labeled by an oracle.

Our SEAL framework comprises three main components, namely, a graph embedding network $G(\cdot)$, a pool tuning (PT), and a discriminator network $D(\cdot)$. As shown in Fig. 1, its workflow operates as follows.

1) Taking graph $\mathcal{G}$ as input, the graph embedding network $G(\cdot)$ encodes both the labeled and unlabeled nodes into low-dimensional, latent node representations, $H_L$ and $H_U$, respectively, with the aim to characterize their class attributes and fool the discriminator $D(\cdot)$ simultaneously.
2) The latent node representations and their prediction probabilities are then passed to PT. PT picks a portion of nodes with high prediction certainty from the unlabeled pool $U$ and moves them to the labeled pool $L$. The two tuned pools are named pseudo labeled (p-labeled) pool $L^+$ and pseudo unlabeled (p-unlabeled) pool $U^-$, respectively. Correspondingly, their latent representations are denoted as $H_{L^+}$ and $H_{U^-}$.
3) The discriminator network $D(\cdot)$ takes $H_{L^+}$ and $H_{U^-}$ as input and maps them into a latent space to generate multiple outputs. These outputs not only produce the probabilities of nodes belonging to $K$ classes but also generate a scoring function to quantify the informativeness of unlabeled nodes with respect to the existing labeled data. The unlabeled node with the highest score from the p-unlabeled pool $U^-$ is selected to be labeled. After that, the original labeled pool $L$ and unlabeled pool $U$ are updated and reinput to $G(\cdot)$.

Using this adversarial learning approach, the graph embedding network and the discriminator network form a closed loop to collaborate and reinforce each other.

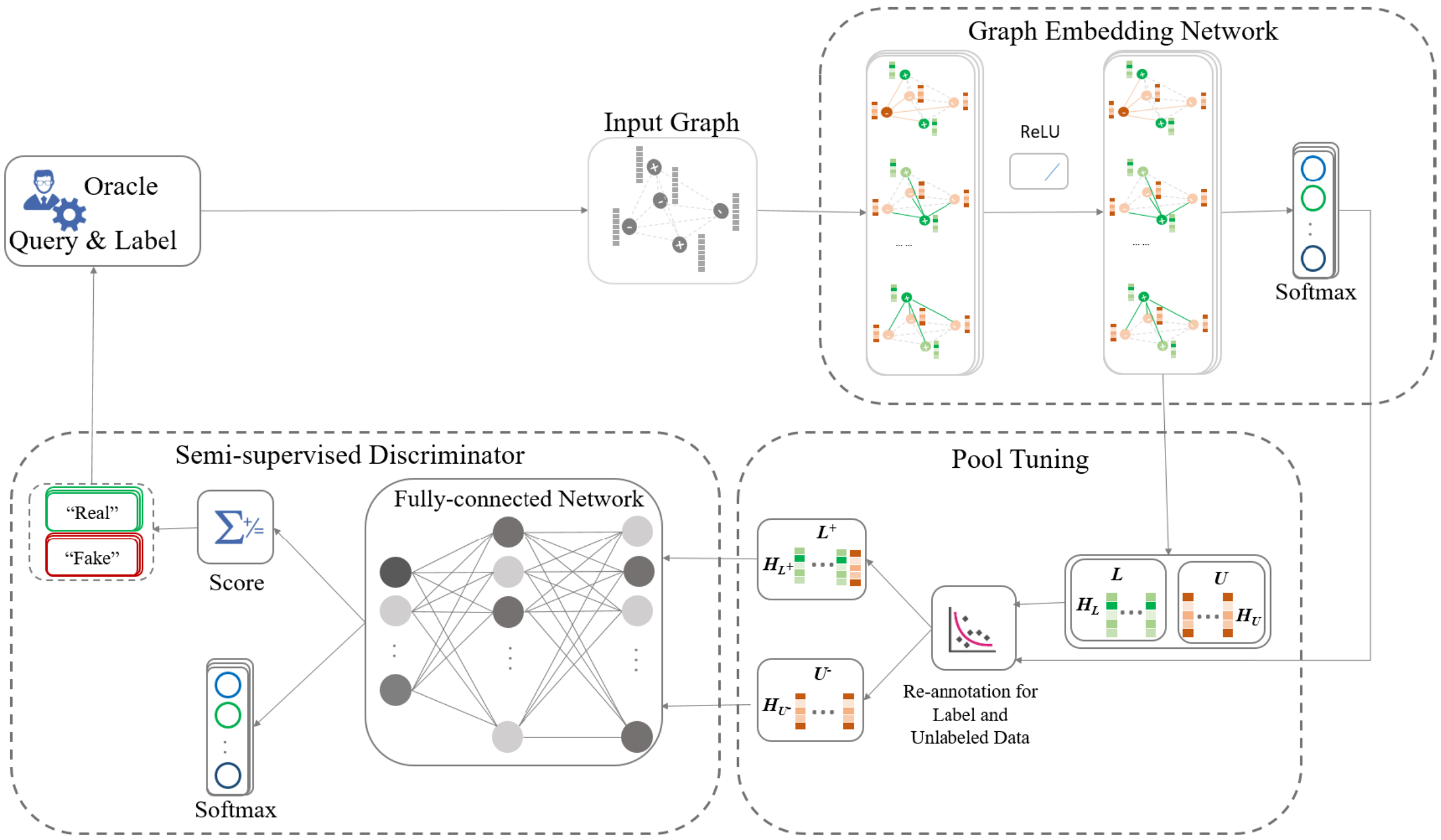

Fig. 1. SEAL framework is composed of three main components: a graph embedding network, a pool tuning (PT), and a semisupervised discriminator network. The graph embedding network encodes both labeled and unlabeled nodes into low-dimensional, latent node representations $H_L$ and $H_U$, respectively, and tricks the discriminator to regard all nodes with latent representations as already labeled. PT tunes the labeled and unlabeled pool $L$ and $U$ to generate p-labeled pool $L^+$ and p-unlabeled pool $U^-$, according to the uncertainty of prediction. The semisupervised discriminator learns to distinguish the unlabeled from labeled nodes and generates an informative score in a common latent space, which enables the selection of the most informative node from p-unlabeled pool $U^-$ to be labeled by an oracle.

### *B. Graph Embedding Network*

Our graph embedding network $G(\cdot)$ encodes all nodes into a low-dimensional, latent embedding space with better representation and discrimination power for learning an AL classifier. For this purpose, we use a specifically modified GCN as the classifier and the representation learner in the SEAL framework. On the one hand, $G(\cdot)$ takes the same responsibility as GCN to learn the latent feature representations of both labeled and unlabeled nodes by using both graph structure and node attributes. The latent node representations are used to learn a classifier for prediction in a latent space. Thus, the cross-entropy loss as (3) is incorporated into our loss function for predicting the labels of graph nodes.

On the other hand, $G(\cdot)$ is expected to provide guidance for the downstream discriminator to improve its capability of measuring the divergence of unlabeled nodes with respect to the existing labeled nodes. Inspired by the adversarial idea of GAN, the loss derived from distinguishing the labeled and unlabeled nodes by the discriminator $D(\cdot)$ is backpropagated to $G(\cdot)$ to subsequently reinforce the generalization and discrimination capability of the discriminator. In this way, $G(\cdot)$ intends to fool the discriminator $D(\cdot)$ to believe that all nodes are from the labeled pool, while the discriminator tries to learn how to differentiate the unlabeled from the existing labeled nodes. Concretely, instead of having the discriminator $D(\cdot)$ output a score "1" for both labeled and unlabeled nodes as the cross-entropy loss does, we employ feature matching method to rectify their feature distributions, which was empirically proved to be more effective in situations where traditional cross-entropy-based supervised methods are volatile [41]. This method is aimed at minimizing the mean discrepancy of feature distributions between p-labeled and p-unlabeled data obtained from the intermediate layer of the discriminator so that the p-unlabeled data can match the statistics of p-labeled data. We assume that $G^{(m)}$ and $D^{(n)}$ represent the hidden representation of the $m$-layer of graph embedding network and $n$-layer of the discriminator, respectively.

Finally, the loss function of our graph embedding network is formulated as

$$J_G = \|\mathbb{E}_{\mathbf{x}\sim L^+} D^{(n)}(G^{(m)}(\mathbf{x})) - \mathbb{E}_{\mathbf{x}\sim U^-} D^{(n)}(G^{(m)}(\mathbf{x}))\|^2 + J_{\text{GCN}} \quad (5)$$

where the first term measures the mean feature discrepancies of nodes in the p-labeled pool $L^+$ and the p-unlabeled pool $U^-$, backpropagated from the discriminator, and the second term is the cross-entropy loss that is calculated using the existing labeled data as (3).

By minimizing this loss function, $G(\cdot)$ tries to not only minimize the classification loss of predicting the labels of nodes but also to force the distribution $\mathbf{x} \sim \mathcal{P}_{U^-}$ to approximate the distribution of $\mathbf{x} \sim \mathcal{P}_{L^+}$. As the training proceeds, by descending mean feature discrepancies sent from the discriminator, $G(\cdot)$ is able to push the two distributions closer. The indistinguishable distributions, reversely, drive $D(\cdot)$ to improve its classification ability by descending its labeled-unlabeled classification errors. Thus, $G(\cdot)$ and $D(\cdot)$ form a GAN-like framework, where $G(\cdot)$ embeds both the labeled and unlabeled nodes into a common latent space with the uniform distribution to fool the discriminator, whereas $D(\cdot)$ intends to distinguish the unlabeled nodes from the labeled pool as accurately as possible. By appropriately parameterizing and optimizing $G(\cdot)$ and $D(\cdot)$, the adversarial process can iteratively strengthen the discriminator with high generalization and discrimination capability.

### C. Pool Tuning

Our objective is to select unlabeled nodes that can provide auxiliary information that has not been captured by the classifier from the labeled data yet. After several epoches of training, it can be assumed that the information of labeled data has been sufficiently acquired by GCN. As for the unlabeled data, many unlabeled nodes, especially those for which the current classifier can predict with high certainty, carry similar information that the current classifier has already captured, due to local dependences between the neighboring nodes. Thus, we tune the distribution of the labeled and unlabeled data according to the uncertainty predicted by the current classifier.

According to the GCN's estimated probability distribution, we reannotate a portion of unlabeled nodes with high predicting confidence as the pseudo labeled data and exclude them from the unlabeled pool. The tuned labeled pool and unlabeled pool are denoted as $L^+$ and $U^-$, respectively. We use a threshold $\delta$, whose value will be empirically determined, to decide which unlabeled nodes should be moved to the labeled pool. Specifically, for any node whose predicting probability on any class exceeds the threshold $\delta$, it would be reannotated and put into the labeled pool $L^+$ using the following equations:

$$L^+ = L \cup \{\mathbf{x}_i \in U | P(\hat{\mathbf{y}}|\mathbf{x}_i) > \delta\} \tag{6}$$

$$U^- = \{\mathbf{x}_i \in U | P(\hat{\mathbf{y}}|\mathbf{x}_i) <= \delta\} \tag{7}$$

where $\hat{\mathbf{y}}$ denotes the most probable class that $\mathbf{x}_i$ belongs to. $L^+$ and $U^-$ are then fed to our discriminator that decides the most informative node to be selected from $U^-$.

### D. Semisupervised Adversarial Learning

Following PT, we design a discriminator network $D(\cdot)$ that approximates a divergence measure to gauge the discrepancy between the p-unlabeled and p-labeled data distribution. In other words, the discriminator tries to tell apart p-unlabeled nodes from p-labeled nodes by minimizing an appropriate loss function.

Instead of using a simple cross-entropy-based binary discriminator, we design a semisupervised discriminator $D(\cdot)$ that outputs $K+1$ probabilities, where $K$ probabilities correspond to probabilities of the node belonging to the $K$ specific classes, and one probability corresponds to the probability of the node being from the unlabeled pool [41]. It has the advantage of being able to distinguish the p-unlabeled from p-labeled nodes but also being capable to predict the probabilities of nodes belonging to specific classes. Generally speaking, the appropriately trained discriminator could lie its decision boundary between data manifolds of different classes, which would in turn improves the generalization performance of the discriminator [42]. Thus, in our AL problem, this objective naturally achieves a good tradeoff between the exploitation that finds the most informative nodes to improve the prediction task and the exploration in the latent feature space.

For a $K$-class problem, we assume that $D(\cdot)$ takes $G^{(m)}(\mathbf{x})$ as input and outputs $(K+1)$-dimensional logits $l_1(\mathbf{x}), l_2(\mathbf{x}), \ldots, l_{K+1}(\mathbf{x})$. These logits, by applying a softmax function, are then turned into class probabilities $P(\mathbf{y} = j \mid \mathbf{x})$ $(j \in 1, 2, \ldots, K+1)$, of which $P(\mathbf{y} = K+1 \mid \mathbf{x})$ represents the probability of $\mathbf{x}$ being unlabeled. Thus, the loss function of this discriminator can be formulated as follows:

$$J_D = \alpha \cdot J_{\text{sup}} + J_{\text{unsup}} \tag{8}$$

$$J_{\text{sup}} = -\mathbb{E}_{\mathbf{x} \sim L} \log P(\mathbf{y} \mid \mathbf{x}, \mathbf{y} < K+1) \tag{9}$$

$$J_{\text{unsup}} = -\{\mathbb{E}_{\mathbf{x} \sim L^+} \log(D(G^{(m)}(\mathbf{x}))) + \mathbb{E}_{\mathbf{x} \sim U^-} \log(1 - D(G^{(m)}(\mathbf{x})))\} \tag{10}$$

$$D(G^{(m)}(\mathbf{x})) = 1 - P(\mathbf{y} = K+1 \mid \mathbf{x}) \tag{11}$$

where $J_{\text{sup}}$ and $J_{\text{unsup}}$ denote the supervised loss and unsupervised loss, respectively. The two components are balanced by a hyperparameter $\alpha$. $J_{\text{sup}}$ is calculated with only the original labeled data $L$ using cross entropy, whereas $J_{\text{unsup}}$ is calculated using both p-labeled nodes $L^+$ and p-unlabeled nodes $U^-$ via the adversarial training method. $D(G^{(m)}(\mathbf{x}))$ represents the likelihood of node $\mathbf{x}$ being p-labeled. The optimal solution to minimizing $J_{\text{sup}}$ and $J_{\text{unsup}}$ is to have $e^{l_j(\mathbf{x})} = c(\mathbf{x})p(\mathbf{y} = j, \mathbf{x}), \forall j < K+1$ and $e^{l_{K+1}(\mathbf{x})} = c(\mathbf{x})p(\mathbf{y} = K+1, \mathbf{x})$ for some undetermined scaling function $c(\mathbf{x})$. In other words, this means that a perfect solution to minimizing $J_{\text{unsup}}$ is also perfect to minimizing $J_{\text{sup}}$. Thus, the consistence of $J_{\text{sup}}$ and $J_{\text{unsup}}$ could guarantee that the optimization of $J_{\text{unsup}}$ also helps improve the supervised performance [43]. As such, we expect to better estimate the optimal solution by minimizing the two loss functions jointly.

Furthermore, because $(K+1)$-output classification tends to have the overparameterized problem, we adopt the following strategy. Given that subtracting a term $f(\mathbf{x})$ would not change the softmax distribution, we fix the last output logit $l_{K+1}(\mathbf{x})$ as zero by operating the following equation:

$$\hat{l}_j(\mathbf{x}) = l_j(\mathbf{x}) - f(\mathbf{x}) \quad \forall j \leq K+1. \tag{12}$$

Therefore, $J_{\text{sup}}$ is recast into a standard supervised loss function with $K$ classes. The probability of nodes being labeled is thus given by

$$D(\mathbf{x}) = \frac{\sum_{k=1}^{K} e^{\hat{l}_k(\mathbf{x})}}{\sum_{k=1}^{K} e^{\hat{l}_k(\mathbf{x})} + 1}. \tag{13}$$

### E. Active Scoring

For AL, we define an informativeness measure based on the divergence between the p-labeled nodes and p-unlabeled nodes. Intuitively, the more divergent an unlabeled node is from the existing labeled data, the more likely it would contribute useful information to the current classifier. As described in Section IV-D, the output of the discriminator can provide a divergence measure. Thus, we devise an active scoring function as

$$\text{div}(\mathbf{x}_{U^-}, L^+) = 1 - D(\mathbf{x}_{U^-}). \tag{14}$$

Intuitively, the higher the score is, the more informative the p-unlabeled node is with respect to the existing labeled data. Consequently, we select node $x^*$ from the p-unlabeled pool $U^-$ such that div $(x^*, L^+)$ is maximized and query its label.

**Algorithm 1:** SEAL Model Training

**Input:** Graph $\mathcal{G}(\mathcal{V}, \mathcal{E}, \mathcal{X})$, node sets $L, U$, labeling budget $B$, pre-training epoches $n_p$, training epoches $n_G$ and $n_D$ for $G(\cdot)$ and $D(\cdot)$

**Output:** a set of selected nodes $L_t$

1 Initialize the parameter of $G(\cdot)$ and $D(\cdot)$ network;
2 **while** *not converged* **do**
3     **for** $t_G = 0; t_G < n_G; t_G = t_G + 1$ **do**
4         Update $G(\cdot)$ by descending gradients of Eq.(5);
5     Tune and generate the candidate pools $L_t^+$ and $U_t^-$ based on Eq.(6) and Eq.(7);
6     **for** $t_D = 0; t_D < n_D; t_D = t_D + 1$ **do**
7         Update $D(\cdot)$ by descending gradients of Eq.(8);
8     **if** $t > n_p$ *and* $|L_t| - |L| < B$ **then**
9         Calculate scores for nodes in $U_t^-$ using Eq.(14);
10        Select node $\mathbf{x}^* \leftarrow \text{argmax}\, div(\mathbf{x}_{U_t^-}, L_t^+)$;
11        Update pools: $L_t \leftarrow L_t \cup \mathbf{x}^*$, $U_t \leftarrow U_t \setminus \mathbf{x}^*$;
12 **return** a set of selected nodes $L_t$;

### F. Model Training and Complexity Analysis

The model training of SEAL is given in Algorithm 1. In the main training loop, we iteratively train the graph embedding network $G(\cdot)$ and the discriminator $D(\cdot)$ (lines 3–7) and then proceed with the instance selection process (lines 8–11). The major computational cost of Algorithm 1 lies in training $G(\cdot)$ and $D(\cdot)$. The computational complexity of a two-layer GCN is linear with the number of edges $|\mathcal{E}|$, i.e., $\mathcal{O}(|\mathcal{E}|HKM)$, where $M$ denotes the dimension of node features, $H$ denotes the number of hidden layer units, and $K$ denotes the number of classes. The computational complexity of the three-layer discriminator is linear with the number of nodes, i.e., $\mathcal{O}(NKH_1H_2|M'|)$, where $|M'|$ denotes the dimension of the input node representations, and $H_1$ and $H_2$ denote the number of the two hidden layer units, respectively. Therefore, in the training process, the overall computational complexity of SEAL is $\mathcal{O}(|\mathcal{E}|HKM + NKH_1H_2|M'|)$, which is linear with the number of edges $|\mathcal{E}|$ and the number of nodes $N$.

### G. Discussion: Differences From GAN

Our SEAL framework is designed as a GAN-like architecture. However, its training objective is essentially different from a regular GAN. As mentioned in Section III-C, a regular GAN generally aims to obtain a perfect generator, and it finally converges at the state where the generator exactly recovers the training data distribution while predicting probability of the discriminator equals to 0.5 everywhere. In contrast, our objective is to obtain a strong discriminator that can measure the divergence between the labeled and unlabeled data distribution with high confidence to enable active instance selection.

TABLE II
STATISTICS OF DATA SETS

| Dataset | Nodes | Edges | Classes | Features | $\lvert L_{init}\rvert / \lvert L_{max}\rvert$ |
|---|---|---|---|---|---|
| Citeseer | 3327 | 4732 | 6 | 3703 | 24 / 120 |
| Cora | 2708 | 5429 | 7 | 1433 | 28 / 140 |
| DBLP | 18447 | 91052 | 4 | 2476 | 16 / 80 |
| Pubmed | 19717 | 44338 | 3 | 500 | 12 / 60 |

If the $G(\cdot)$ achieves the exact match between the distribution of $x \sim \mathcal{P}_{U^-}$ and $x \sim \mathcal{P}_{L^+}$, for any optimal solution $\mathcal{S}$ to the supervised loss $J_{\text{sup}}$, there exists an optimal solution $\mathcal{S}^*$ to the semisupervised $(K + 1)$-class objective $J_D$ such that $\mathcal{S}$ and $\mathcal{S}^*$ share the same generalization error. Therefore, under the semisupervised setting in SEAL, a perfect $G(\cdot)$ that can exactly match the two distributions of $x \sim \mathcal{P}_{L^+}$ and $x \sim \mathcal{P}_{U^-}$ would not be able to improve the generalization capability of the discriminator over the supervised setting. Consequently, a weaker $G(\cdot)$ would be necessary to guarantee a stronger discriminator. Thus, it is required to appropriately optimize (5) and (8) using an alternating optimization switching between the updates to $G(\cdot)$ and $D(\cdot)$. While this optimization is not guaranteed to converge, empirically, it provides us a strong discriminator if $G(\cdot)$ and $D(\cdot)$ are well balanced. This is consistent with the existing findings in [44].

## V. EXPERIMENTAL ANALYSIS

To validate the effectiveness of our SEAL framework, a series of experiments is conducted on node classification tasks under a transductive, pool-based AL setting: Given the initial labeled nodes and a certain labeling budget, unlabeled nodes are iteratively selected to label and train a classifier, whose performance is tested from different perspectives.

### A. Data Sets

Four benchmark citation networks, including Citeseer,[1] Cora[1], Pubmed[1] [45], and DBLP,[2] [46] are used in our experiments. In these networks, each node represents a document with a certain label and each edge represents the citation links between two documents. We treat these networks as undirected and unweighted graphs, and each node is characterized by a sparse bag-of-words feature vector according to word occurrence. DBLP is a subgraph of the DBLP bibliographic network, including publications from four research areas. Cora, Citeseer, and DBLP are used to evaluate the classification-related performance, and Pubmed is used to test the computational complexity. Details of the four data sets are summarized in Table II.

At the beginning of the training process, only $|L_{\text{init}}|$ labels are accessible, and as AL algorithms proceed, one unlabeled node is selected to label at each iteration. A maximum number of $|L_{\text{max}}|$ labels can be queried. In our implementation, for each data set, we start with four labeled nodes per class

[1] https://linqs.soe.ucsc.edu/data
[2] https://aminer.org/citation

$|L_{\text{init}}| = 4 \times K$, as the initial state, and the labeling budget $B = 20 \times K - |L_{\text{init}}|$, where $K$ denotes the number of classes.

### B. Experimental Setup

Our experiments closely follow the settings as in [1] and [19]. For each data set, we randomly choose 1000 nodes for testing, 500 nodes for validation. To quantify the performance difference induced only by different AL query strategies, instead of randomly sampling 500 nodes as the validation set for each run of experiments, we generate ten different validation sets by randomly sampling from the nontesting unlabeled pool and repeat experiments for ten times on each validation set. This setting is designed to ensure that the same unlabeled pool is used when running different AL query strategies. Finally, the reported results are averaged over 10 (validation sets) $\times 10$ (initial labeled sets).

We utilize a two-layer GCN network, and its hidden layer has 16 units. ReLU and $L2$ regularization ($5 \times 10^{-4}$) are applied for the first layer only. It is optimized using Adam, where the learning rate is 0.005 for the instance selection period and 0.01 for the subsequent node prediction period, and the dropout rate is 0.5. The discriminator is a three-layer fully connected neural network with $(128, 128, K)$ units, respectively, and each layer is followed by Leaky ReLU activation. It is also optimized using Adam, where the learning rate is 0.01 and the dropout rate is 0.5. Before the AL query process starts, the whole network is pretrained for 300 epoches, i.e., $n_p = 300$ in Algorithm 1, to ensure that $G(\cdot)$ has adequate representation learning capacity and $D(\cdot)$ has adequate discrimination ability.

### C. Baselines

We compare our SEAL framework against four state-of-the-art AL methods, with details as follows.

1) *AGE [19] and ANRMAB [20]:* They are two state-of-the-art methods that combine GCN with classic AL strategies, via a linear combination of three AL query strategies (graph centrality, information density, and uncertainty sampling). ANRMAB improves AGE by dynamically adjusting the weights of different strategies based on the MAB reward. They differ from SEAL in terms of different AL query strategies designed, which are used to validate the efficacy of SEAL's unified AL query strategy.
2) *ALFNET [12]:* It is a traditional AL strategy that uses ICA and QBC ensemble to make instance selection. This method is used to evaluate the advantages of GNN-based AL methods over traditional graph-based AL methods. In our settings, we adapt it as a transductive semisupervised version and allow it to select only one node at each iteration for fair comparison.
3) *GCN-Random [1]:* It uses GCN as the classifier but randomly chooses one unlabeled node to query its label.

To assess the importance of different aspects of SEAL, we also compare with four variants of SEAL via ablation studies.

1) *SEAL-ad:* It is a variant of SEAL with adversarial learning obliterated. Specifically, it changes $G(\cdot)$'s loss function in (5) as $J_G = J_{\text{GCN}}$. $D(\cdot)$ is still used to discriminate unlabeled from labeled nodes with loss function $J_D$ as (8), but loss of $D(\cdot)$ is not backpropagated to $G(\cdot)$. It is designed to validate the effectiveness of the adversarial learning mechanism.
2) *SEAL-fm:* This method is a variant of SEAL to test the effectiveness of feature matching loss for generator $G(\cdot)$. Instead of minimizing feature matching loss as (5), it maximizes the log-likelihood of both labeled and unlabeled nodes to confuse $D(\cdot)$, which is given by

$$J_G = -\mathbb{E}_{\mathbf{x} \sim L \bigcup U} D(G^{(m)}(\mathbf{x})) + J_{\text{GCN}}. \tag{15}$$

3) *SEAL-sal:* This method is another variant of SEAL that uses a cross-entropy based binary p-labeled/p-unlabeled discriminator. It is equivalent to setting $\alpha$ as 0 in (8), while other parameters remain the same as with SEAL.
4) *SEAL-pt:* This method differs from SEAL in which it removes the operation of PT, and the generated nodes representations are directly sent to $D(\cdot)$. This is equivalent to setting $\delta$ to 1 in (6) and (7). Other parameters remain the same with SEAL.

For classification, ALFNET uses the ICA with logistic regression as the base classifier. All other algorithms use GCN as the classifier for fair comparison.

### D. Overall Performance Comparison

We use Micro-F1 and Macro-F1 scores as the evaluation criteria to validate the node classification performance. Table III compares the performance of different algorithms on Citeseer, Cora, and DBLP. For SEAL, we set parameters $\alpha$ and $\delta$ to 0.6 for this experiment. Overall, as can be seen, SEAL exhibits evident advantages over other baselines by designing a new AL query strategy on a unified scoring space. In terms of Micro-F1 score, SEAL improves upon ANRMAB by a margin of 1.3%, 1.2%, and 2.4%, respectively, on the three data sets. Similar results can also be seen on the Macro-F1 score results. This is in accordance with our expectation that the discriminator allows us to further exploit the GNN-generated representations in a new latent space, where dependences among these latent representations can be better captured to support instance selection. Furthermore, our mechanism could select nodes with the least redundant information. The two reasons lead to superior classification performance to weighted combination methods, such as AGE and ANRMAB. It is also worth noting that ALFNET performs consistently worse than any other GNN-based method, including GCN-Random. This proves the advantageous representation power of GNN over traditional ICA-based methods.

### E. Ablation Study

We also conduct an ablation study to assess the importance of different components of SEAL. Our analysis is reported in the bottom section of Table III.

Overall, SEAL has stable performance gains over its degraded counterparts, due to the contributions of different

TABLE III

MICRO-F1 AND MACRO-F1 PERFORMANCE COMPARISON WITH $L_{\max}$ LABELED NODES FOR TRAINING

| Method | Citeseer | | Cora | | DBLP | |
|---|---|---|---|---|---|---|
| | Micro-F1 | Macro-F1 | Micro-F1 | Macro-F1 | Micro-F1 | Macro-F1 |
| GCN-Random | 0.685 | 0.615 | 0.812 | 0.794 | 0.720 | 0.702 |
| AGE | 0.717 | 0.666 | 0.813 | 0.800 | 0.772 | 0.719 |
| ANRMAB | 0.721 | 0.672 | 0.819 | 0.807 | 0.778 | 0.739 |
| ALFNET | 0.650 | 0.613 | 0.765 | 0.750 | 0.656 | 0.628 |
| SEAL-ad | 0.712 | 0.660 | 0.819 | 0.809 | 0.757 | 0.706 |
| SEAL-fm | 0.721 | 0.665 | 0.825 | 0.816 | 0.760 | 0.705 |
| SEAL-sal | 0.717 | 0.662 | 0.817 | 0.806 | 0.762 | 0.716 |
| SEAL-pt | 0.733 | 0.675 | 0.829 | 0.815 | 0.782 | 0.711 |
| SEAL | **0.734** | **0.676** | **0.831** | **0.822** | **0.802** | **0.747** |

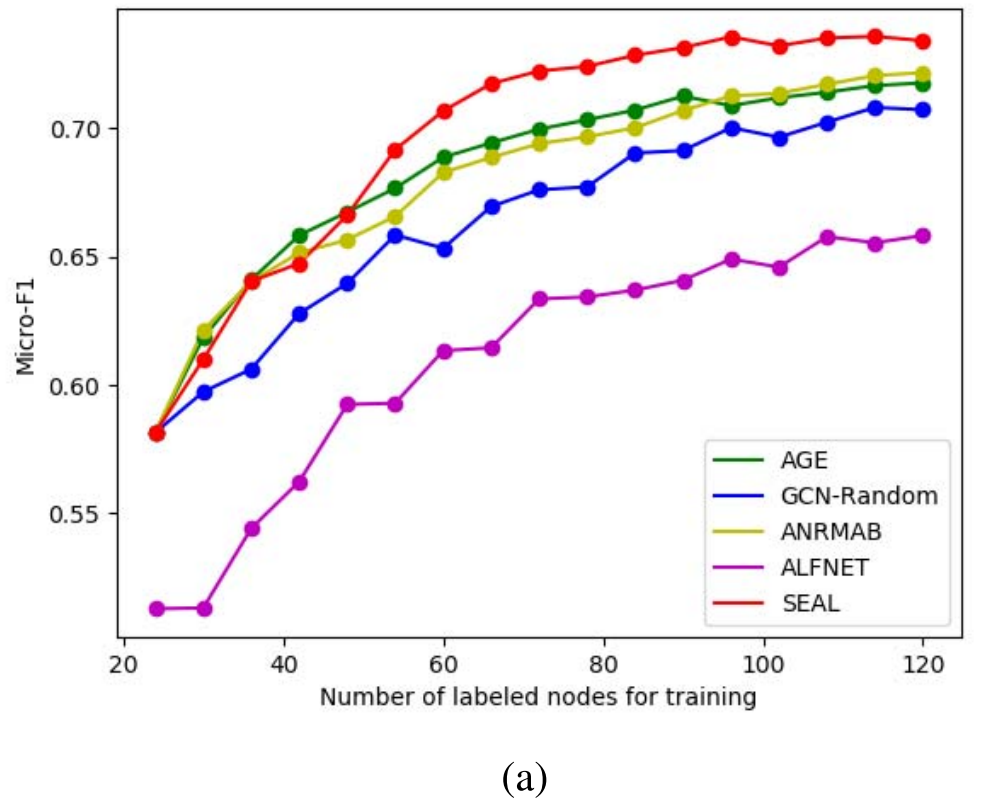

(a)

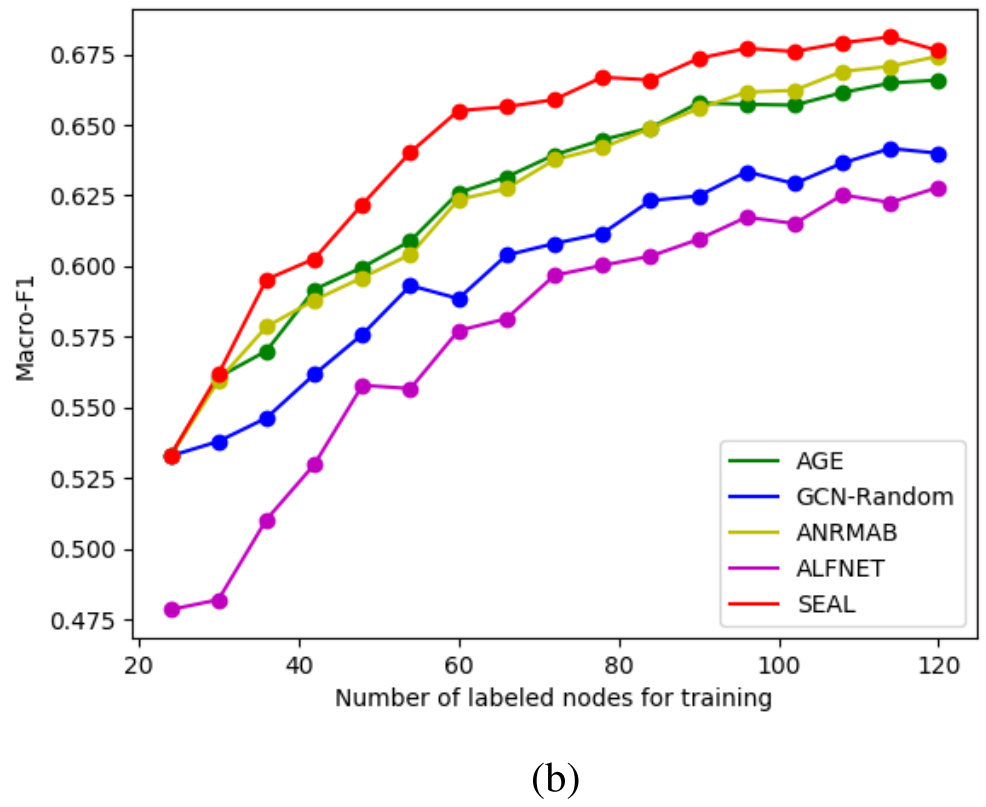

(b)

Fig. 2. Performance comparison with respect to different labeling budgets on Citeseer. (a) Micro-F1. (b) Macro-F1.

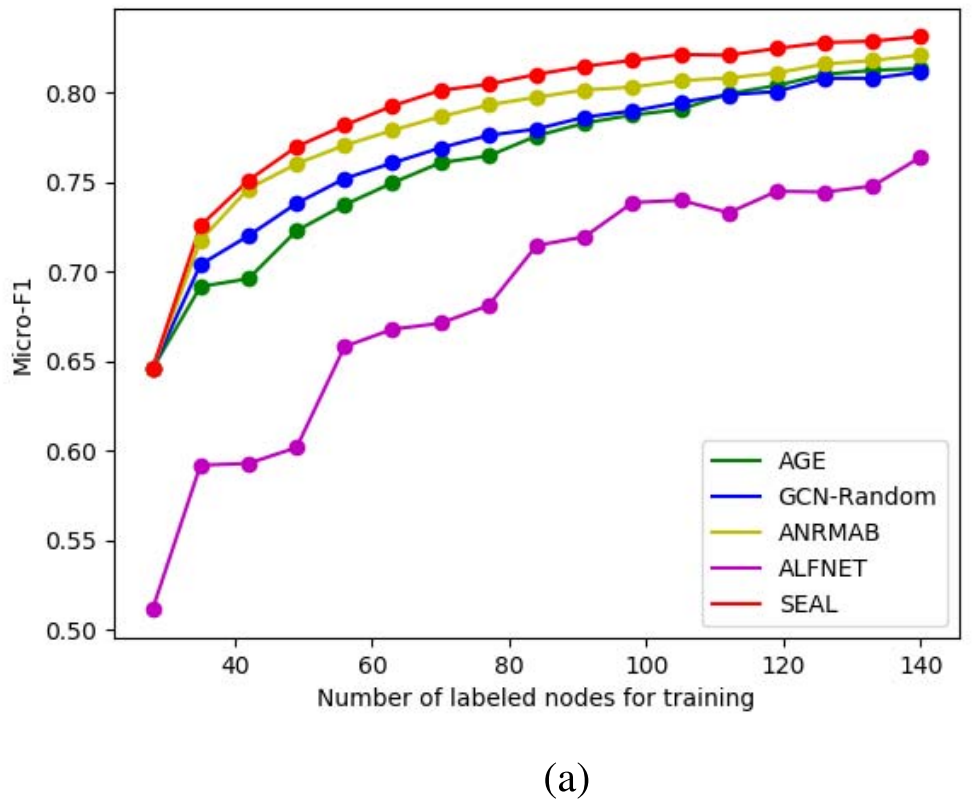

(a)

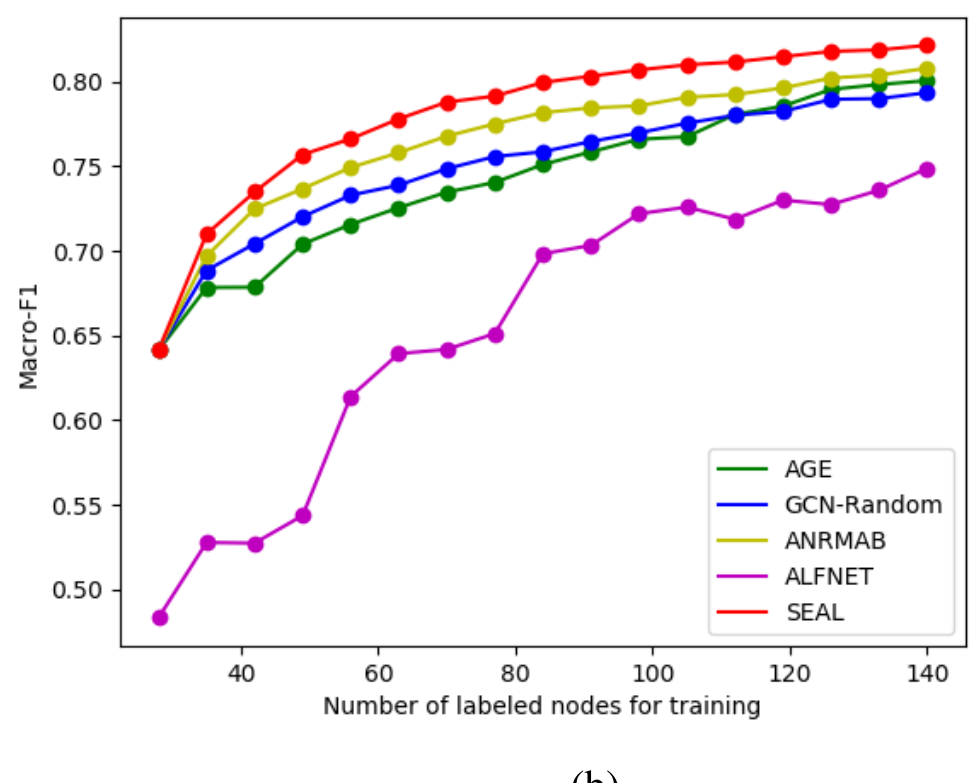

(b)

Fig. 3. Performance comparison with respect to different labeling budgets on Cora. (a) Micro-F1. (b) Macro-F1.

components. The adversarial leaning mechanism (SEAL-ad) allows $G(\cdot)$ and $D(\cdot)$ to reinforce each other and thus boost the classification ability of $D(\cdot)$. This enables us to select the most informative samples for improving the overall performance. Feature matching (SEAL-fm), as a method for alleviating overfitting and mode collapse problems in GAN [41], effectively stabilizes the training of SEAL. PT (SEAL-pt), redistributing labeled and unlabeled nodes, enables $D(\cdot)$ to better distinguish unlabeled nodes and to reduce the search space of unlabeled candidates. Semisupervised adversarial learning mechanism (SEAL-sal) allows $D(\cdot)$ to measure the divergence between the labeled and unlabeled nodes. Effectiveness of PT and semisupervised adversarial learning mechanism will be further analyzed carefully in Section V-G and V-H.

### *F. Performance Comparison of Different Labeling Budgets*

Figs. 2–4 compare the classification performance of different methods with respect to different labeling budgets on Citeseer, Cora, and DBLP. Although all methods have an overall upward trend as the number of labeled nodes increases, SEAL offers the steepest improvement slopes with remarkable gains over other baselines. Taking Fig. 2(a) as an example, SEAL reaches 72.0% of classification accuracy with only 66 labeled nodes, whereas ANRMAB reaches the similar accuracy until obtaining 120 labeled nodes. This indicates that SEAL achieves similar classification accuracy with much fewer nodes labeled, which again proves the effectiveness of our AL strategy.

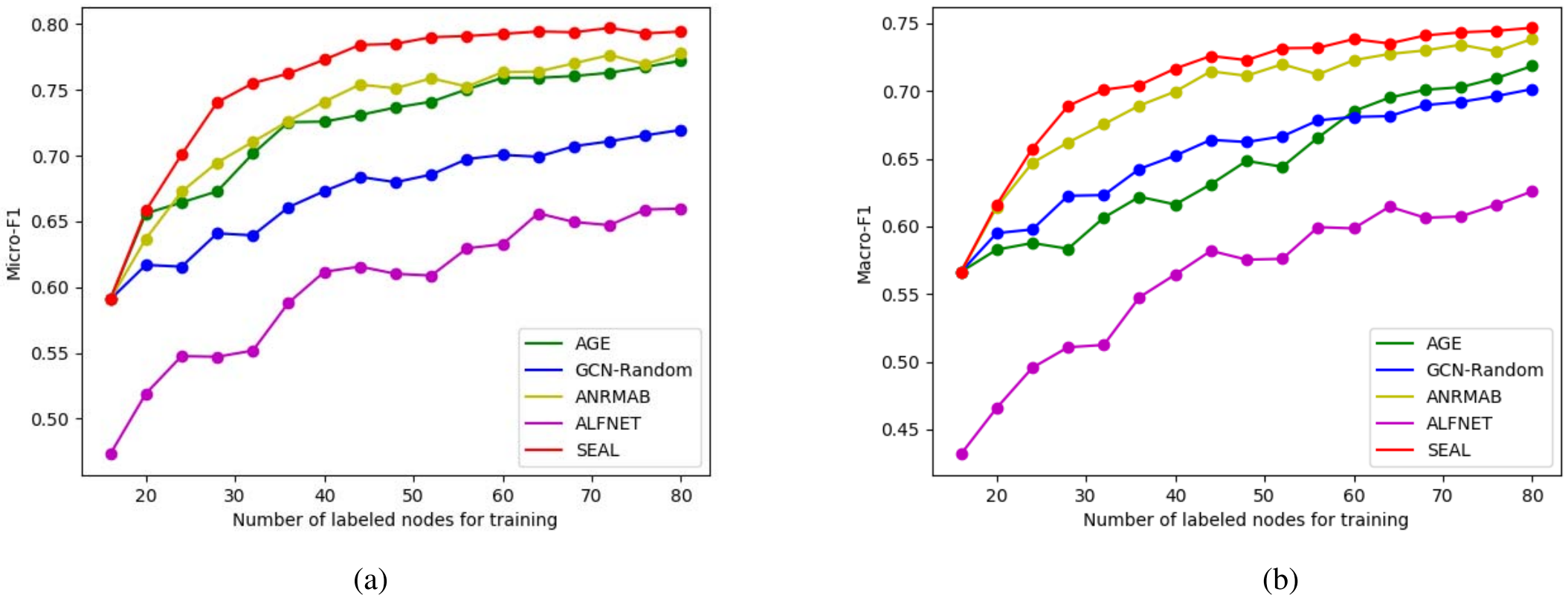

Fig. 4. Performance comparison with respect to different labeling budgets on DBLP. (a) Micro-F1. (b) Macro-F1.

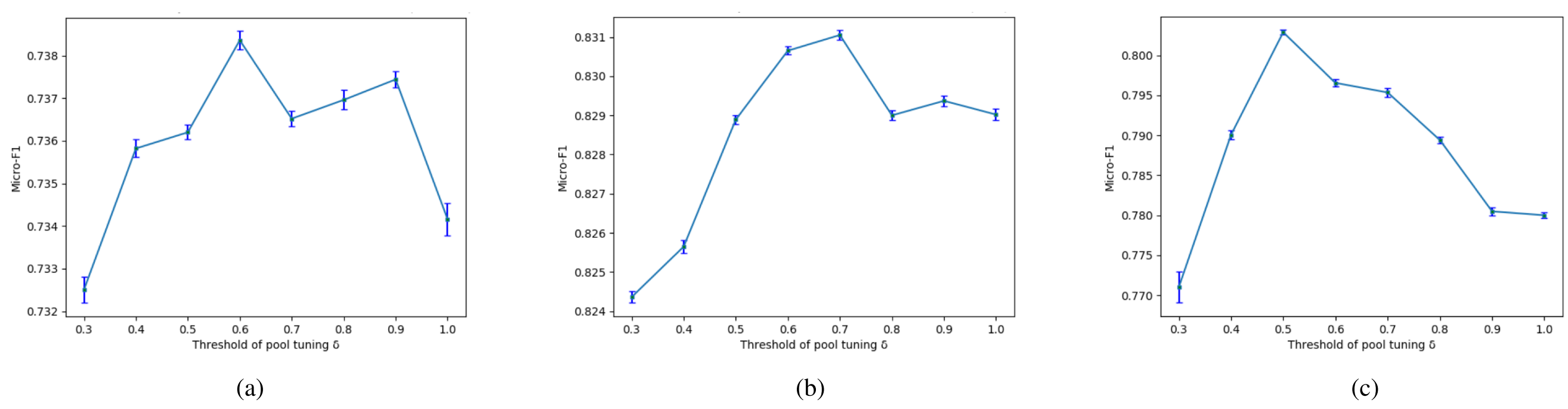

Fig. 5. Comparison of Micro-F1 with respect to varying $\delta$ values on (a) Citeseer, (b) Cora, and (c) DBLP.

### *G. Effectiveness Study on PT*

Fig. 5 shows the performance changes with varying thresholds $\delta$ for PT , where a similar trend can be observed on Citeseer, Cora, and DBLP. As can be seen, SEAL achieves the best performance when $\delta$ is equal to 0.6 and degrades with varying speeds as $\delta$ further increases. The standard variances are plotted as a vertical line at each point, which are calculated over 100 repeating tests on each threshold. The longer the vertical line, the larger the standard variance. PT picks a bunch of unlabeled nodes carrying similar information with already labeled nodes as pseudolabeled nodes. The redistributed node sets make it easier for $D(\cdot)$ to find the most distinct unlabeled nodes. Moreover, it helps avoid the potential overfitting caused by extremely imbalanced labeled/unlabeled samples, thus leading to more stable performance.

### *H. Effectiveness Study on the SAL*

Fig. 6 shows the sensitivity of SEAL with respect to $\alpha$ that balances the tradeoff between $J_{\text{sup}}$ and $J_{\text{unsup}}$ in (8). Taking DBLP in Fig. 6(c) as an example, when $\alpha$ is zero, $J_D$ degrades as an unsupervised loss function (i.e., $J_D = J_{\text{unsup}}$), where only 76.2% of predictions are correctly made. Then, the performance fluctuates with the increase of $\alpha$, during which it reaches the peak at the point around $\alpha = 0.6$. An appropriate $\alpha$ allows $D(\cdot)$ to be aware of differentiation between different classes when measuring the similarity between the p-unlabeled and p-labeled nodes. This awareness alleviates the risk of selecting abnormal nodes as it often occurs in uncertainty sampling methods, thereby resulting in better classification accuracy.

### *I. Training Time Comparison and Convergence Rate*

We also conduct experiments to compare the training time (in seconds) of three GCN-based AL methods, SEAL, AGE, and ANRMAB. All methods are implemented in tensorflow on a Linux system with Intel Xeon CPU E5-2690 at 3.4 GHz*8 and 32-GB memory. We compare their training time on Pubmed, where the number of nodes increases from 1000 to 19 000 with an increment of 2000. We record the total training time for selecting 48 unlabeled nodes for labeling, and the results are shown in Fig. 7(a). As we can see, ANRMAB and AGE take almost the same amount of training time, which is less than SEAL at the beginning, when the network size is small. However, as network size increases, the training time of ANRMAB/AGE rapidly grows at a nonlinear rate, which is several times more than that of SEAL. This is because the frequent sorting operation in ANRMAB/AGE incurs a computational overhead of $\mathcal{O}(N^2)$, whereas SEAL maintains a linear growth rate with respect to a number of nodes $N$. This indicates that SEAL's computational overhead for instance selection is acceptable and it is reasonably efficient on large graphs.

Fig. 7(b) shows the changes in training loss $J_G$ on Pubmed as the training proceeds. In the beginning, as new unlabeled nodes start to be added into the labeled set, the training loss exhibits some fluctuations. This is probably due to that conspicuously distinct information carried by the newly added nodes disturbs parameter updates of the classifier trained by the existing labeled data. As more nodes are labeled, the training loss quickly decreases and stabilizes at a lower level. The overall trend of loss $J_G$ exhibits good convergence.

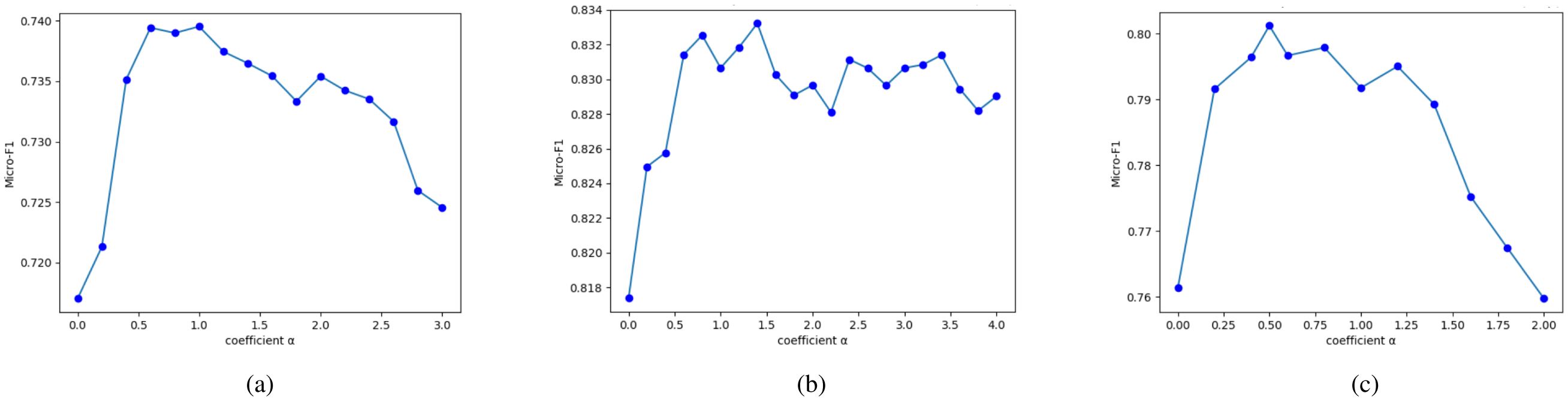

Fig. 6. Comparison of Micro-F1 scores with respect to varying $\alpha$ values on (a) Citeseer, (b) Cora, and (c) DBLP.

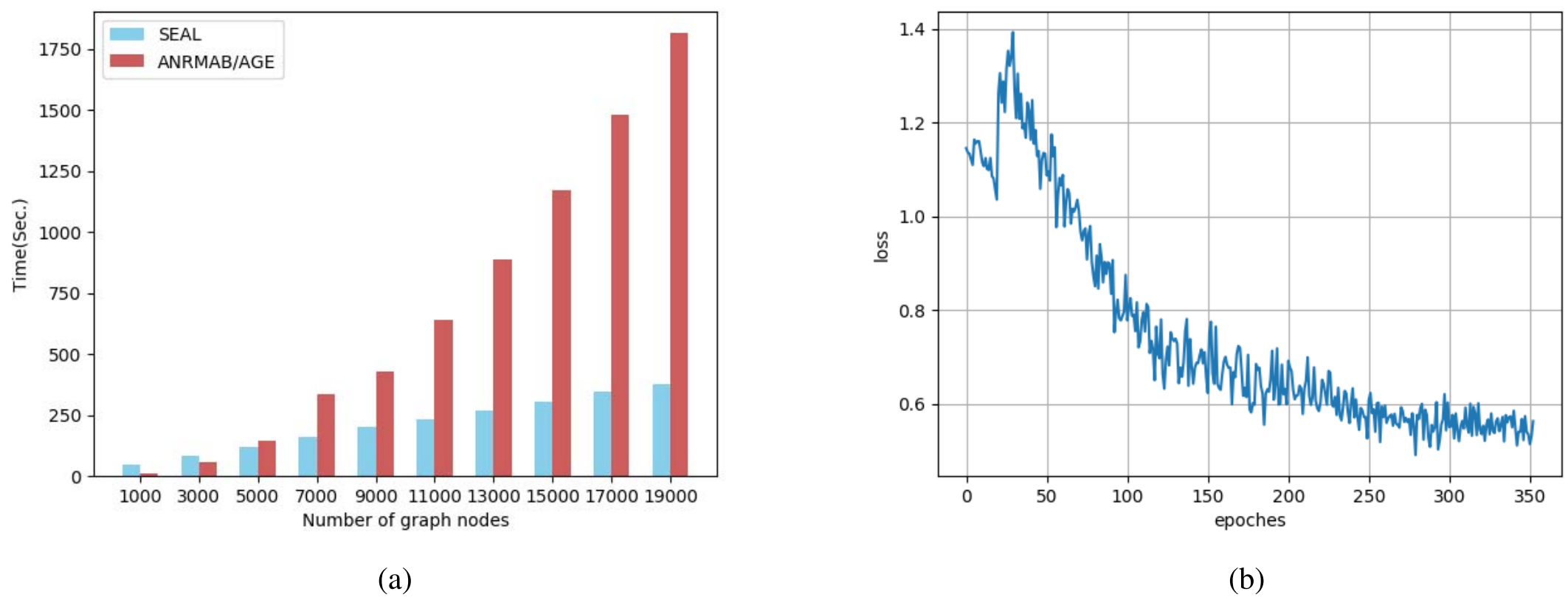

Fig. 7. Training time and convergence analysis on Pubmed. (a) Comparison of training time. (b) Convergence of training loss.

## VI. Conclusion and Future Work

In this article, we addressed an AL problem on attributed graphs, which requires to take both graph structure and node attributes into account. We argued that the existing AL frameworks are ineffective in: 1) they use a naive weighted combination of different AL strategies to select the nodes to label and 2) they treat the learning of graph embedding and the AL query engine as two separate and independent processes, leading to limited AL performance gains. To fill the research gap, we proposed a novel semisupervised AL framework called SEAL, which fully exploits the representation power of GNNs and devises a novel AL query strategy for graphs. SEAL comprises two adversarial components; a graph embedding network is trained to embed both the labeled and unlabeled nodes into a common latent space and to trick the discriminator to believe that all nodes are from the labeled pool, while a discriminator network learns how to tell them apart using a semisupervised structure with multiple outputs. The divergence score, produced by the discriminator, serves as the informativeness measure to select the most useful node to be labeled by the oracle. The two adversarial components form a closed loop to mutually and simultaneously reinforce each other toward enhancing the AL performance. Extensive experiments and ablation studies proved that the SEAL renders remarkable performance gains compared with state-of-the-art AL methods on node classification tasks. In this work, we use a modified GCN as the classifier and the representation learner for node classification. For future work, we would like to adapt and test our proposed AL framework on other network architectures and graph-related tasks.

## References

[1] T. N. Kipf and M. Welling, “Semi-supervised classification with graph convolutional networks,” in *Proc. ICLR*, 2017, pp. 1–13.

[2] S. Sun and D. R. Hardoon, “Active learning with extremely sparse labeled examples,” *Neurocomputing*, vol. 73, nos. 16–18, pp. 2980–2988, Oct. 2010.

[3] M.-W. Chang, L.-A. Ratinov, N. Rizzolo, and D. Roth, “Learning and inference with constraints,” in *Proc. AAAI*, 2008, pp. 1513–1518.

[4] D. D. Lewis and J. Catlett, “Heterogeneous uncertainty sampling for supervised learning,” in *Machine Learning Proceedings*. Amsterdam, The Netherlands: Elsevier, 1994, pp. 148–156.

[5] N. Roy and A. McCallum, “Toward optimal active learning through monte carlo estimation of error reduction,” in *Proc. ICML*, vol. 2001, pp. 441–448.

[6] H. S. Seung, M. Opper, and H. Sompolinsky, “Query by committee,” in *Proc. 5th Annu. Workshop Comput. Learn. Theory (COLT)*, 1992, pp. 287–294.

[7] A. K. McCallumzy and K. Nigamy, “Employing EM and pool-based active learning for text classification,” in *Proc. ICML*, 1998, pp. 359–367.

[8] D. D. Lewis and W. A. Gale, “A sequential algorithm for training text classifiers,” in *Proc. SIGIR*. Dublin, Republic of Ireland: Springer, 1994, pp. 3–12.

[9] M. Ji and J. Han, "A variance minimization criterion to active learning on graphs," in *Proc. AISTATS*, 2012, pp. 556–564.

[10] Q. Gu and J. Han, "Towards active learning on graphs: An error bound minimization approach," in *Proc. IEEE 12th Int. Conf. Data Mining*, Dec. 2012, pp. 882–887.

[11] Y. Ma, R. Garnett, and J. Schneider, "$\sigma$-optimality for active learning on Gaussian random fields," in *Proc. NeurIPS*, 2013, pp. 2751–2759.

[12] M. Bilgic, L. Mihalkova, and L. Getoor, "Active learning for networked data," in *Proc. ICML*, 2010, pp. 79–86.

[13] M. Fang, J. Yin, X. Zhu, and C. Zhang, "Active class discovery and learning for networked data," in *Proc. SIAM Int. Conf. Data Mining*, May 2013, pp. 315–323.

[14] D. Berberidis and G. B. Giannakis, "Active sampling for graph-aware classification," in *Proc. IEEE Global Conf. Signal Inf. Process. (GlobalSIP)*, Nov. 2017, pp. 648–652.

[15] P. Veličković, G. Cucurull, A. Casanova, A. Romero, P. Liò, and Y. Bengio, "Graph attention networks," 2017, *arXiv:1710.10903*. [Online]. Available: http://arxiv.org/abs/1710.10903

[16] R. Hong, Y. He, L. Wu, Y. Ge, and X. Wu, "Deep attributed network embedding by preserving structure and attribute information," *IEEE Trans. Syst., Man, Cybern. Syst.*, early access, Mar. 1, 2019, doi: 10.1109/TSMC.2019.2897152.

[17] Q. Lu and L. Getoor, "Link-based classification," in *Proc. ICML*, 2003, pp. 496–503.

[18] B. Perozzi, R. Al-Rfou, and S. Skiena, "Deepwalk: Online learning of social representations," in *Proc. 20th ACM SIGKDD Int. Conf. Knowl. Discovery Data Mining*, 2014, pp. 701–710.

[19] H. Cai, V. W. Zheng, and K. Chen-Chuan Chang, "Active learning for graph embedding," 2017, *arXiv:1705.05085*. [Online]. Available: http://arxiv.org/abs/1705.05085

[20] L. Gao, H. Yang, C. Zhou, J. Wu, S. Pan, and Y. Hu, "Active discriminative network representation learning," in *Proc. 27th Int. Joint Conf. Artif. Intell.*, Jul. 2018, pp. 2142–2148.

[21] B. Settles, M. Craven, and S. Ray, "Multiple-instance active learning," in *Proc. NeurIPS*, 2008, pp. 1289–1296.

[22] A. J. Joshi, F. Porikli, and N. P. Papanikolopoulos, "Scalable active learning for multiclass image classification," *IEEE Trans. Pattern Anal. Mach. Intell.*, vol. 34, no. 11, pp. 2259–2273, Nov. 2012.

[23] S. Geman, E. Bienenstock, and R. Doursat, "Neural networks and the bias/variance dilemma," *Neural Comput.*, vol. 4, no. 1, pp. 1–58, Jan. 1992.

[24] D. J. C. MacKay, "Information-based objective functions for active data selection," *Neural Comput.*, vol. 4, no. 4, pp. 590–604, Jul. 1992.

[25] B. Settles and M. Craven, "An analysis of active learning strategies for sequence labeling tasks," in *Proc. Conf. Empirical Methods Natural Lang. Process. (EMNLP)*, 2008, pp. 1070–1079.

[26] A. Fujii, T. Tokunaga, K. Inui, and H. Tanaka, "Selective sampling for example-based word sense disambiguation," *Comput. Linguist.*, vol. 24, no. 4, pp. 573–597, Dec. 1998.

[27] B. Settles, "Active learning literature survey," Dept. Comput. Sci., Univ. Wisconsin-Madison, Madison, WI, USA, Tech. Rep. 1648, 2009.

[28] C. C. Aggarwal, *Data Classification: Algorithms and Applications*. Boca Raton, FL, USA: CRC Press, 2014.

[29] L. Shi, Y. Zhao, and J. Tang, "Batch mode active learning for networked data," *ACM Trans. Intell. Syst. Technol.*, vol. 3, no. 2, pp. 1–25, 2012.

[30] A. Guillory and J. A. Bilmes, "Active semi-supervised learning using submodular functions," 2012, *arXiv:1202.3726*. [Online]. Available: http://arxiv.org/abs/1202.3726

[31] A. Guillory and J. A. Bilmes, "Label selection on graphs," in *Proc. NeurIPS*, 2009, pp. 691–699.

[32] J. Long, J. Yin, W. Zhao, and E. Zhu, "Graph-based active learning based on label propagation," in *Proc. MDAI*. Catalonia, Spain: Springer, 2008, pp. 179–190.

[33] W. Zhao, J. Long, E. Zhu, and Y. Liu, "A scalable algorithm for graph-based active learning," in *Proc. FAW*. Changsha, China: Springer, 2008, pp. 311–322.

[34] X. Zhu, J. Lafferty, and Z. Ghahramani, "Combining active learning and semi-supervised learning using Gaussian fields and harmonic functions," in *ICML workshop*, vol. 3, 2003, pp. 1–8.

[35] D. Berberidis and G. B. Giannakis, "Data-adaptive active sampling for efficient graph-cognizant classification," *IEEE Trans. Signal Process.*, vol. 66, no. 19, pp. 5167–5179, Oct. 2018.

[36] L. Shi, Y. Zhao, and J. Tang, "Combining link and content for collective active learning," in *Proc. 19th ACM Int. Conf. Inf. Knowl. Manage. (CIKM)*, 2010, pp. 1829–1832.

[37] Y. Fu, X. Zhu, and B. Li, "A survey on instance selection for active learning," *Knowl. Inf. Syst.*, vol. 35, no. 2, pp. 249–283, May 2013.

[38] I. Goodfellow, *et al.*, "Generative adversarial nets," in *Proc. NeurIPS*, 2014, pp. 2672–2680.

[39] Y. Deng, K. Chen, Y. Shen, and H. Jin, "Adversarial active learning for sequences labeling and generation," in *Proc. 27th Int. Joint Conf. Artif. Intell.*, Jul. 2018, pp. 4012–4018.

[40] S. Sinha, S. Ebrahimi, and T. Darrell, "Variational adversarial active learning," in *Proc. IEEE/CVF Int. Conf. Comput. Vis. (ICCV)*, Oct. 2019, pp. 5972–5981.

[41] T. Salimans, I. Goodfellow, W. Zaremba, V. Cheung, A. Radford, and X. Chen, "Improved techniques for training GANs," in *Proc. NeurIPS*, 2016, pp. 2234–2242.

[42] W. Li, Z. Wang, J. Li, J. Polson, W. Speier, and C. Arnold, "Semi-supervised learning based on generative adversarial network: A comparison between good GAN and bad GAN approach," in *Proc. CVPR Workshop*, 2019, pp. 1–11.

[43] I. Sutskever, R. Jozefowicz, K. Gregor, D. Rezende, T. Lillicrap, and O. Vinyals, "Towards principled unsupervised learning," 2015, *arXiv:1511.06440*. [Online]. Available: http://arxiv.org/abs/1511.06440

[44] Z. Dai, Z. Yang, F. Yang, W. W. Cohen, and R. R. Salakhutdinov, "Good semi-supervised learning that requires a bad GAN," in *Proc. NeurIPS*, 2017, pp. 6510–6520.

[45] P. Sen, G. Namata, M. Bilgic, L. Getoor, B. Galligher, and T. Eliassi-Rad, "Collective classification in network data," *AI Mag.*, vol. 29, no. 3, p. 93, Sep. 2008.

[46] D. Zhang, J. Yin, X. Zhu, and C. Zhang, "Attributed network embedding via subspace discovery," *Data Mining Knowl. Discovery*, vol. 33, no. 6, pp. 1953–1980, Nov. 2019.

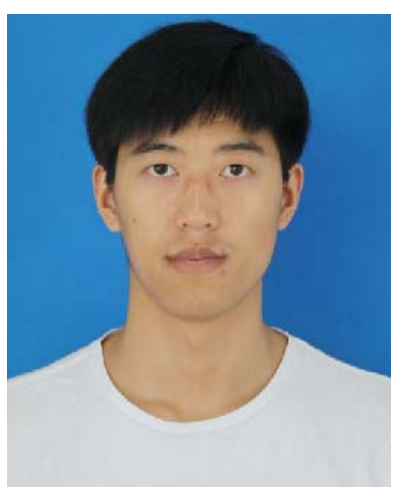

**Yayong Li** received the master's degree from the University of Electronic Science and Technology of China, Chengdu, China, in 2018. He is currently pursuing the Ph.D. degree with the Centre for Artificial Intelligence, University of Technology Sydney, Sydney, NSW, Australia.

His research interests include data mining and machine learning, especially on graph analysis and applications.

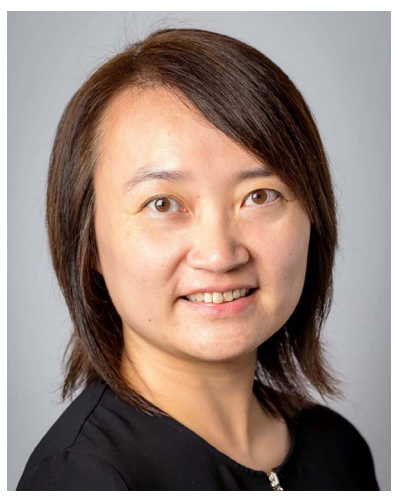

**Jie Yin** (Member, IEEE) received the Ph.D. degree in computer science from The Hong Kong University of Science and Technology, Hong Kong, in 2006.

She is currently a Senior Lecturer with the Discipline of Business Analytics, The University of Sydney, Sydney, NSW, Australia. Her research interests include data mining and machine learning, with a focus on graph analysis, text mining, social network analysis, and interpretable AI.

Dr. Yin was recognized as the AI 2000 AAAI/IJCAI Most Influential Scholar Award Honorable Mention in 2020. She was the Co-Chair of the International Workshop on Social Web for Disaster Management from 2015 to 2018 and a Guest Editor of Special Issue of the IEEE Intelligent Systems on AI for Disaster Management and Resilience.

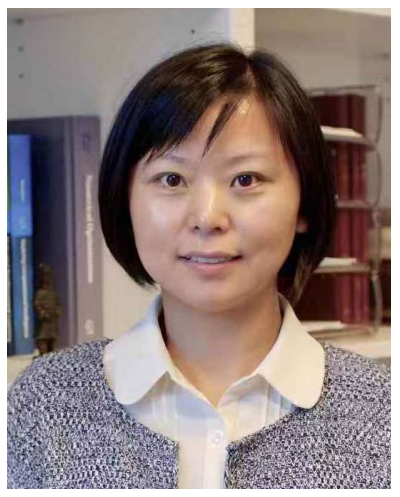

**Ling Chen** (Senior Member, IEEE) received the Ph.D. degree from Nanyang Technological University, Singapore, in 2008.

She is currently an Associate Professor with the Centre for Artificial Intelligence, University of Technology Sydney, Sydney, NSW, Australia. Her papers appear in major conferences and journals, including SIGKDD, IJCAI, the IEEE TRANSACTIONS ON NEURAL NETWORKS AND LEARNING SYSTEMS, and the IEEE TRANSACTIONS ON KNOWLEDGE AND DATA ENGINEERING. Her research interests include data mining and machine learning, especially on structured data such as graph data and spatiotemporal data. She also works on social networks and social media analysis and applications.